# Unmanned Aerial Vehicle (UAV) Data-Driven Modeling Software with Integrated 9-Axis IMU-GPS Sensor Fusion and Data Filtering Algorithm

Azfar Azdi Arfakhsyad
*Department of Electrical Engineering and Information Technology*
*Universitas Gadjah Mada*
Yogyakarta, Indonesia
azfar.azdi.arfakhsyad@mail.ugm.ac.id

Aufa Nasywa Rahman
*Department of Electrical Engineering and Information Technology*
*Universitas Gadjah Mada*
Yogyakarta, Indonesia
aufa.nas2003@mail.ugm.ac.id

Larasati Kinanti
*Department of Mechanical and Industrial Engineering*
*Universitas Gadjah Mada*
Yogyakarta, Indonesia
larasati.kinanti@mail.ugm.ac.id

Ahmad Ataka Awwalur Rizqi
*Department of Electrical Engineering and Information Technology*
*Universitas Gadjah Mada*
Yogyakarta, Indonesia
ahmad.ataka.ar@ugm.ac.id

Hannan Nur Muhammad
*Department of Electrical Engineering and Information Technology*
*Universitas Gadjah Mada*
Yogyakarta, Indonesia
hannan.nur.muhammad@mail.ugm.ac.id

*Abstract*—Unmanned Aerial Vehicles (UAV) have emerged as versatile platforms, driving the demand for accurate modeling to support developmental testing. This paper proposes data-driven modeling software for UAV. Emphasizes the utilization of cost-effective sensors to obtain orientation and location data subsequently processed through the application of data filtering algorithms and sensor fusion techniques to improve the data quality to make a precise model visualization on the software. UAV's orientation is obtained using processed Inertial Measurement Unit (IMU) data and represented using Quaternion Representation to avoid the gimbal lock problem. The UAV's location is determined by combining data from the Global Positioning System (GPS), which provides stable geographic coordinates but slower data update frequency, and the accelerometer, which has higher data update frequency but integrating it to get position data is unstable due to its accumulative error. By combining data from these two sensors, the software is able to calculate and continuously update the UAV's real-time position during its flight operations. The result shows that the software effectively renders UAV orientation and position with high degree of accuracy and fluidity.

## I. INTRODUCTION

Unmanned Aerial Vehicles (UAV) have rapidly evolved as a versatile platform for various applications [1]. The increasing demand for UAV development to solve complex environments necessitates raising the need to develop accurate and reliable simulation models that faithfully represent the dynamic behavior of the UAV. An accurate simulation model of UAV that has been tested allows developers to perform cost-effective analysis and evaluation while also validating the performance of UAV under real-world scenarios. Analyzing UAV behavior under various real-world scenarios is expensive, time-consuming, and requires extensive planning. Moreover, it is very risky because there is a chance that UAV fails due to plenty of reasons with no data collected.

Traditional analysis often requires human intervention and many simplifications that do not fully capture the complexity of real-world scenarios. Traditional modeling, such as using a camera to capture UAV orientation, only gave us visual representation. As the result, we cannot simulate what is happening on the UAV using numerical data. On the other hand, simulation-based analysis gave us visual representation and numerical data that can be used to calculate and analyze further.

Accurate modeling of UAV's dynamics is challenging due to their complexity and sensitivity to external factors such as noise. In this paper, Sensor fusion techniques are used to overcome those issues. Sensor fusion integrates data from multiple sensors, providing more comprehensive understanding of the UAV orientation and position. The most used sensor fusion algorithm is complementary filter [2]-[4] and Kalman filter [5]-[7]. Complementary filter acts as a sensor fusion technique implemented in this paper since it does not use as many resources as Kalman filter [8], and it could perform very well if the parameters are well configured [9]. The UAV model can be more reliable and accurate by fusing Inertial Measurement Unit (IMU) data and position information from sensors such as the Global Positioning System (GPS). Quaternion representation is used to represent the UAV orientation. One of the key benefits of using quaternion is their ability to avoid the gimbal lock problem found in Euler angles representation [10]. Gimbal lock describes a condition in which one of the rotation axes becomes aligned with another, leading to a loss of a degree of freedom in the representation [11]. On the other hand, position data is obtained by combining accelerometer and GPS data. The accelerometer data, which captures the acceleration of the UAV, is processed through a series of steps to extract position-related information.

Double integration is employed to derive velocity and subsequently integrate it again to obtain position estimates. Filtering techniques are applied to reduce error accumulation and improve the accuracy of the integration process. It will be covered later how these filters work. These filters aim to remove noise from the accelerometer data.

This proposed model has the ability to simulate UAV orientation and location from a three-dimensional point of view, having high accuracy and lightweight software. Three-dimensional points of view play a crucial role in simulating UAV behavior, allowing users to monitor UAV orientation based on different points of view. High accuracy is also important since it is achieved using simple and inexpensive sensors. Lastly, the proposed software is lightweight, improving UAV modeling performance.

This paper is organized into six different sections. The first section, "Introduction," highlights the significance of data-driven modeling in the context of UAV. It also provides an overview of the approach employed to develop the data-driven simulation. The second section, "System Overview," comprehensively examines the system's construction. The third section, "Concepts and Algorithm," explains all the concepts and algorithms used in this paper, including the sensor fusion and data filtering algorithms. Moving on to the fourth section, "Modeling," the paper explains how the UAV model is visualized. Furthermore, this section also explains all the features available on the software. The fifth section, "Result," presents the outcomes and findings. Finally, the paper concludes with the sixth section, "Conclusion." The key takeaways are summarized.

## II. SYSTEM OVERVIEW

### A. Overall System

The overall system consists of several key components, including the GY-87 module, three microcontrollers ATmega328, u-blox M8N GPS module, and three nRF24L01 modules. The GY-87 module and u-blox M8N GPS module are crucial parts of the system since it is used to capture all the data. The GY-87 module includes the MPU6050 and HMC5883L. The MPU6050 sensor acts as an accelerometer and gyroscope. The HMC5883L sensor allows the system to detect and measure the magnetic field. Two ATmega328 are responsible for processing data from those sensors and transmitting those data through wireless communication. Data transmitted from those two ATmega328 already undergoes data filtering algorithm apart from complementary filter. All algorithms will be explained in the next section.

Two ATmega328 are directly connected to two different sensors, one from the GY-87 module and the other one from u-blox M8N GPS module. The last ATmega328 acts as a receiver. Those three ATmega328 are connected to nRF24L01. Two nRF24L01 acts as transmitters from two different sensors, and one nRF24L01 acts as a receiver from both sensors. The overall system, as described in the paragraph earlier, is visualized in Fig. 1. This figure represents the components and their connections.

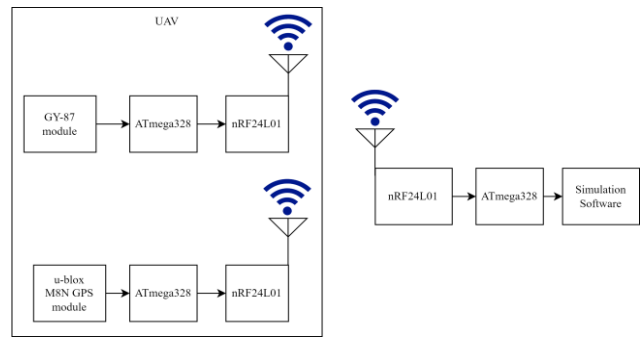

Fig. 1. All components constructing the system.

### B. Orientation Data Processing

Orientation data are acquired from processing accelerometer and gyroscope data from the GY-87 module. Data processing involves several distinct steps. Accelerometer data is processed using Low Pass Filter to reduce noise from high-frequency data since an accelerometer captures slow changes in motion and orientation. Meanwhile, gyroscope data is processed using High Pass Filter. This approach was chosen because gyroscopes are sensitive to low-frequency noise that can cause inaccurate estimation. After both accelerometer and gyroscope data are pre-processed using concepts above, those two different data are combined using complementary filter, resulting in Roll, Pitch, and Yaw representation in Euler Angle. Yaw representation is combined with compass data obtained from the HMC5883L sensor to maximize accuracy. After combining data using the abovementioned concept, Roll, Pitch, and Yaw data are represented in quaternion representation to avoid the gimbal lock problem. The visualization of the concept mentioned above is available in Fig. 2

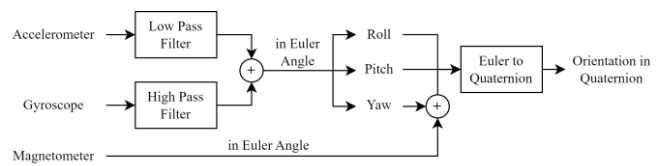

Fig. 2. Steps involved to process orientation data.

### C. Position Data Processing

The processing of position data involves several steps. The accelerometer data, obtained from the MPU6050 sensor, is processed using Butterworth filter. After processing the data, it undergoes double integration. The first integration results in velocity, and the second integration results in displacement. This process essentially integrates the accelerometer data twice to obtain the position data. However, since double integration can accumulate errors over time, combining this information with other data sources is essential for improving accuracy. The processed accelerometer data is combined with GPS data using complementary filter. GPS data provides latitude and longitude measurements. Resulting in improved position estimation. The diagram for gathering and processing position data is visualized in Fig. 3.

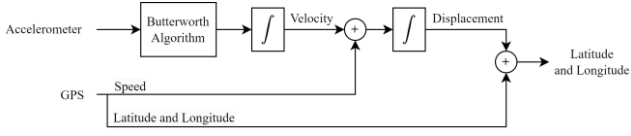

Fig. 3. Steps involved to process position data.

Additional calculations are needed to convert accelerometer data to process with other data sources, in this case, GPS data. The calculations are written as follows

$$V_x = \alpha \int a_x dt + (1-\alpha) \cdot speed_{gps} \cdot \sin(\theta_d) \tag{1}$$

$$V_y = \alpha \int a_y dt + (1-\alpha) \cdot speed_{gps} \cdot \cos(\theta_d) \tag{2}$$

$$\theta_d = \tan^{-1}\left(\frac{\Delta \ell_y}{\Delta \ell_x}\right) \tag{3}$$

$$\Delta \ell_y = \sin(\Delta \theta_{long}) \cos(\Delta \theta'_{lat}) \tag{4}$$

$$\Delta \ell_x = \cos(\theta_{lat}) \sin(\theta'_{lat}) - \sin(\theta_{lat}) \cos(\theta'_{lat}) \cos(\Delta \theta_{long}) \tag{5}$$

$$\Delta \theta_{long} = \theta'_{long} - \theta_{long} \tag{6}$$

$$S_{lat} = \frac{\beta \int V_x dt \cdot 180}{\pi \cdot r_e} + (1-\beta) \cdot \theta_{latgps} \tag{7}$$

$$S_{long} = \frac{\beta \int V_y dt \cdot 180}{\pi \cdot r_e} + (1-\beta) \cdot \theta_{longgps} \tag{8}$$

where some of the terms used above are

$\alpha$ = velocity weight factor
$\beta$ = displacement weight factor
$\theta_d$ = angle from north to the displacement
$\theta'_{lat}, \theta'_{long}$ = next latitude, next longitude
$S_{lat}, S_{long}$ = displacement in latitude after complementary, displacement in longitude after complementary,
$r_e$ = radius of the earth

## III. CONCEPTS AND ALGORITHMS

### A. Quaternion Representation

Quaternion representation is an effective notation used to represent orientation and rotation since it does not suffer from the gimbal lock that occurs in Euler angle representation [10] [12]. The matrix representation in quaternion for angle Roll-Pitch-Yaw ($\psi$, $\theta$, $\phi$) can be derived from several steps. Thus, the rotation quaternion can be written as

$$S\frac{A}{B} = \cos\frac{b}{2} + \sin\frac{b}{2} \tag{9}$$

$$v = v_x + v_y + v_z \tag{10}$$

$$S\frac{A}{B} = \cos\frac{b}{2} - v_x \sin\frac{b}{2} v_y \sin\frac{b}{2} - v_z \sin\frac{b}{2} \tag{11}$$

$$S\frac{A}{B} = S_1 S_2 S_3 S_4 \tag{12}$$

When rotating a vector $V_p$ within frame A by an angle α around vector v, the resulting vector S in frame B can be expressed as follows:

$$V_S = S\frac{A}{B} * V_p * S\frac{\vec{A}}{B} \tag{13}$$

The Hamilton product '*' is used for combining two quaternions, $P_1$ and $P_2$. When applying the Hamilton product to these quaternions, it can be written as follows:

$$P_1 * P_2 = \begin{bmatrix} a_1 a_2 - b_1 b_2 - c_1 c_2 - d_1 d_2 \\ a_1 b_2 + b_1 a_2 - c_1 d_2 - d_1 c_2 \\ a_1 c_2 - b_1 d_2 + c_1 a_2 + d_1 b_2 \\ a_1 d_2 + b_1 b_2 - c_1 b_2 + d_1 a_2 \end{bmatrix} \tag{14}$$

The following representation of the rotational matrix $R\frac{A}{B}$ derived from quaternion $S\frac{A}{B}$

$$R\frac{A}{B} = \begin{bmatrix} -2S_3^2 - 2S_4^2 & 2S_2 S_3 - 2S_4 S_1 & 2S_2 S_4 + 2S_3 S_1 \\ 2S_2 S_3 + 2S_4 S_1 & 1 - 2S_2^2 - 2S_4^2 & 2S_3 S_4 - 2S_2 S_1 \\ 2S_2 S_4 - 2S_3 S_1 & 2S_3 S_4 + 2S_2 S_1 & 1 - 2S_2^2 - 2S_4^2 \end{bmatrix} \tag{15}$$

The angle Roll-Pitch-Yaw ($\psi$, $\theta$, $\phi$) of an UAV can be achieved then,

$$\begin{bmatrix} \psi \\ \theta \\ \phi \end{bmatrix} = \begin{bmatrix} \tan^{-1} 2(S_1 S_2 + S_3 S_4), 1 - 2(S_2^2 + S_4^2) \\ \sin^{-1}(2(S_1 S_2 + S_3 S_4)) \\ \tan^{-1} 2(S_1 S_4 + S_2 S_3), 1 - 2(S_3^2 + S_4^2) \end{bmatrix} \tag{16}$$

### B. Complementary Filter

Complementary filter is used three times. It combines data from an accelerometer with gyroscope, yaw with magnetometer data, and data from an accelerometer with GPS. The complementary filter takes advantage of their individual strengths and compensates for their limitations. In the context of combining accelerometer and gyroscope data, the complementary filter combines those data in the form of an Euler angle. Roll, pitch, and yaw can be obtained.

On the other hand, yaw data denotes UAV movement in orientation or rotational movement. Meanwhile, a magnetometer is also excellent at capturing this aspect. Complementing both data can increase accuracy estimates. This complementing behavior reduces the drift generated from yaw data that has been processed through several steps, allowing noise to accumulate. The concept above is proven, as can be seen in Fig. 7.

In contrast, accelerometer is used to capture short-term and high-frequency motion data, making them suitable for tracking rapid movements and changes in direction. On the other hand, GPS provides global positioning information. GPS provides accurate long-term position estimates but struggles with capturing location in uniform time frequency since it depends on signal availability. The complementary filter calculates the position estimates by blending the accelerometer's measurements with the GPS's position estimates. Due to the accelerometer's drift, the estimation becomes increasingly inaccurate over longer periods. This is where GPS data comes to take a significant effect. The complementary filter uses GPS data to correct the accelerometer's position estimates. The GPS provides a reference point to align the accelerometer's estimates with the actual position, minimizing the accumulated errors.

*C. Butterworth Filter*

Butterworth filter is a type of Infinite-Length Unit Impulse Response (IIR) digital filter. Founded in 1930 by Stephen Butterworth. Butterworth Filter comes close to approximating an ideal filter, as it aims to eliminate unwanted frequencies and maintains uniform sensitivity across the desired passband [13]. Butterworth filter aims to achieve a maximally flat frequency response curve in the passband, resulting in a smooth response [14]. The main purpose of this filter is to minimize any ripples or variations in amplitude across the desired frequency range. This filter cut-off frequency must be configured since it is prone to noise [15]. Mathematically, the general form for the second-order Butterworth Algorithm can be written as [16]

$$B(s) = \frac{\omega_c^2}{(s + \omega_c e^{j(\pi/4)})(s + \omega_c e^{-j(\pi/4)})} \quad (17)$$

$$= \frac{\omega_c^2}{(s^2 + \sqrt{2}\omega_c s + \omega_c^2)} \quad (18)$$

$\omega_c$ represents the cut-off frequency. It is the point at which the filter attenuates the input signal and denotes the complex frequency variable. In general, the numerator parts represent the squared cut-off frequency; meanwhile, the denominator represents a quadratic polynomial in complex frequency. This polynomial usually describes how the filter gain changes with different frequency components in the input signal.

## IV. MODELING

The orientation is visualized in a three-dimensional model, illustrated in two distinct axes. This illustration can be adjusted based on user needs. This visualization is included in Fig. 4. Conversely, position data visualized on the live map can be accessed in real-time using an internet connection. After the software receives the latitude and longitude data, the software automatically plots the UAV on the exact location.

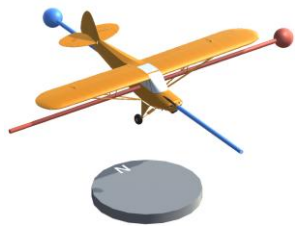

Fig. 4. UAV orientation visualization.

In this paper, software for visualizing the position and orientation of UAV is developed using Unity. The software interface, as depicted in Fig. 5, aims to tell all the necessary information to the user as simply as possible since its primary objective is to facilitate the efficient simulation of UAV. This software includes some features. The first one is a three-dimensional view of the UAV situated in the top left corner. This visualization enables users to observe the UAV orientation through an intuitive and easily interpretable representation. Below that visualization, there is a horizon view and UAV flight panel. Horizon view enables users to view and evaluate UAV roll and pitch behavior. UAV flight panel enables users to view and evaluate distinct data, such as compass, distance to marked location, altitude, speed, roll, pitch, and yaw. A menu bar on the right side of the interface gives users access to various software settings. These settings provide functions such as locating specific positions on the map, viewing latitude and longitude coordinates of the UAV and marked location, adjusting map styles, establishing serial connections to receivers, importing and exporting UAV recorded flight data, and zooming in or out on the maps. Positioned on the bottom of the interface available controls to record flight data, such as start, stop, save, delete recording, and moving recorded data in a specific time.

There are three modes available in this software. The first one is serial mode. The software enters serial mode when the receiver is connected to the software and maintains serial communication with the software. In this mode, IMU and GPS data are received and not recorded. The second mode is recording mode. Recording mode happens when the record button is clicked. The software maintains a serial connection with the receiver alongside recording those data. Those data can then be saved onto Comma Separated Values (CSV) files that can be remodeled. The last mode is simulation mode. Simulation mode happens when recorded data is played back. Recorded IMU and GPS data are read from a CSV file and then shown on the software. In this simulation mode, linear interpolation for position data is implemented. That scenario differs from serial mode, where linear interpolation is not applied. Besides those two different scenarios in which linear interpolation is implemented or not, all sensor fusion and data filtering algorithm explained earlier was implemented on all modes.

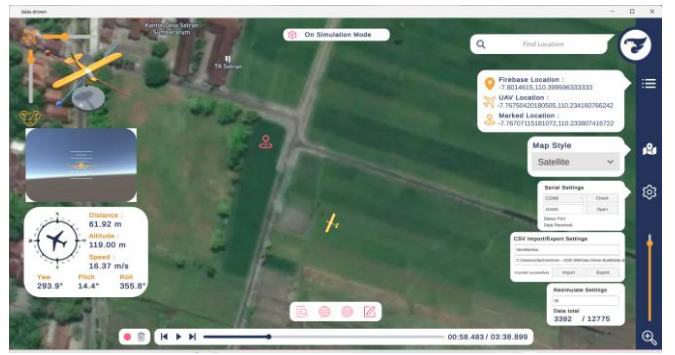

Fig. 5. Software interface.

## V. RESULT

This paper captures an experiment for testing the sensor fusion and data filtering algorithm captured from flying an actual UAV. A fixed-wing UAV model was chosen because fixed-wing serves as benchmark for another UAV models. This implies that when this software is planned to be used using multirotor, tiltrotor, or any other models, it should not pose different challenges. The fixed-wing UAV model can be seen in Fig. 6. The overall system is placed inside the fuselage to ensure that data captured from the sensors represent UAV movement accurately. Data captured from flying an actual UAV for 3 minutes and 38 seconds with 60 FPS is 12774 data from IMU and GPS. All the plots covered later are sampled from the actual flight data.

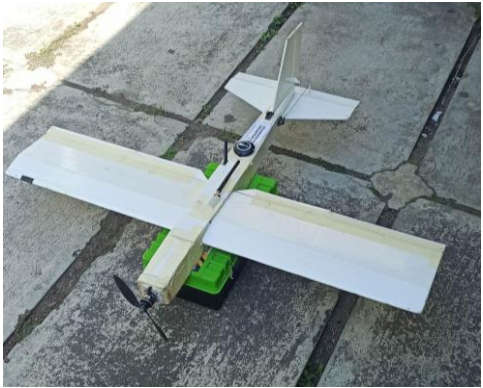

Fig. 6. Fixed wing UAV model that used to capture data.

### A. Orientation

Fig. 7 shows the yaw data comparison. The orange line shows that when integral is performed onto yaw data directly, it will lead to accumulated error. It is shown from the line that slowly moving apart. On the other hand, the green line represents combined data from yaw data and magnetometer data. It can easily be seen that combined data has less accumulated error. That is because combined data is taking magnetometer data into account. The magnetometer acts as a corrector for yaw data that constantly accumulates errors.

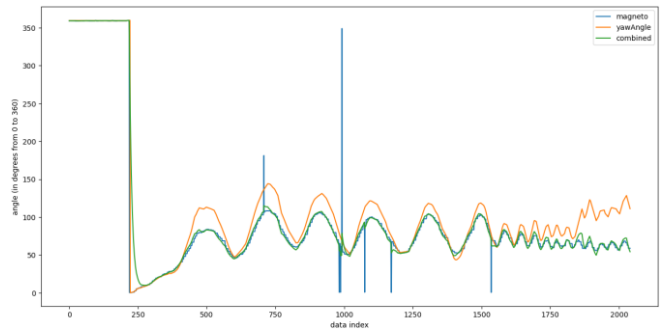

Fig. 7. Yaw data comparison.

Fig. 8 shows the quaternion representation after undergoing different steps to reduce noise that was already explained earlier. There are four different lines in the plot mentioned. There are 'q', 'x', 'y', and 'z' lines. Each of them is represented with a different colored line. The graph might appear unstable and spiky, but it is quite different when it comes to modeling the UAV's orientation with the quaternion data. The graph translates into smooth orientation model due to the nature of quaternion representation.

The 'q' entity represents the quaternion itself and serves as the core of the orientation model. This entity is constructed from a combination of both rotational and angular properties, presented into a single entity. The 'x', 'y', and 'z' entities, on the other hand, represent the individual components of the quaternion. All these entities represent the UAV's dynamics in three different dimensions. The combination of all these components provides a smooth UAV model.

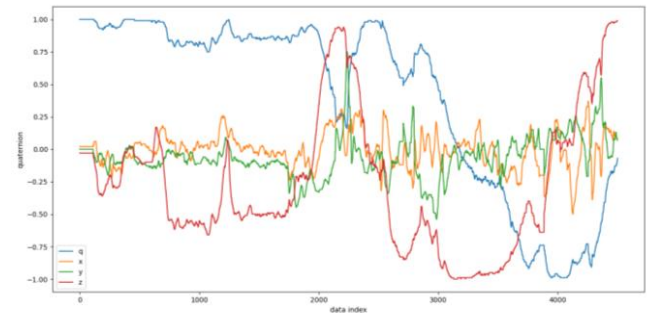

Fig. 8. Quaternion plot

### B. Position

Fig. 9 and Fig. 10 show the acceleration data on the x-axis and y-axis accordingly. The plot consists of three different kinds of data. Which are the original data, data after being processed using the Butterworth filter, and data after being processed using Chebyshev filter. As shown in the plot provided, the original data is very noisy. That is the reason a second-order low-pass filter is used to filter the data. Among many second-order low-pass filters, two of the most popular are Chebyshev and Butterworth filter. Chebyshev and Butterworth filter are compared. Chebyshev and Butterworth filter successfully attenuate noise, denoted by minimized high frequency. However, Chebyshev filter is seen to be more oscillated even though both filters use the same cut-off frequency of 10 Hertz and sampling rate of 1000 data. The frequent oscillation that happens on the Chebyshev filter can cause many problems in the next data processing. When processed into position data, oscillated data can cause a buildup round-off error that happened earlier and a larger error accumulated. Chebyshev filter also uses more resources that are identified by longer processing time. Bigger resource use and longer processing time can cause inaccurate UAV orientation and position estimation. Those two limitations provide a strong baseline to use the Butterworth filter over the Chebyshev filter.

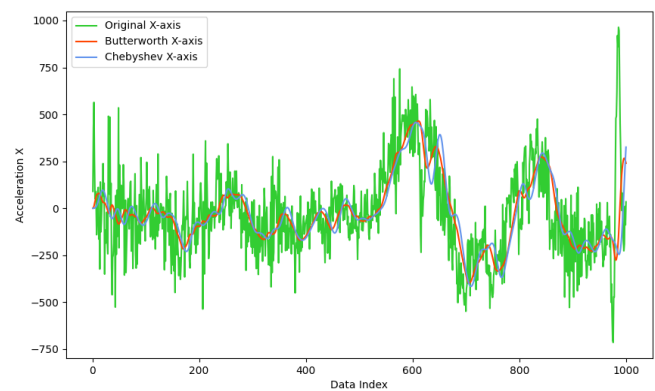

Fig. 9. Acceleration data on x-axis using different filters.

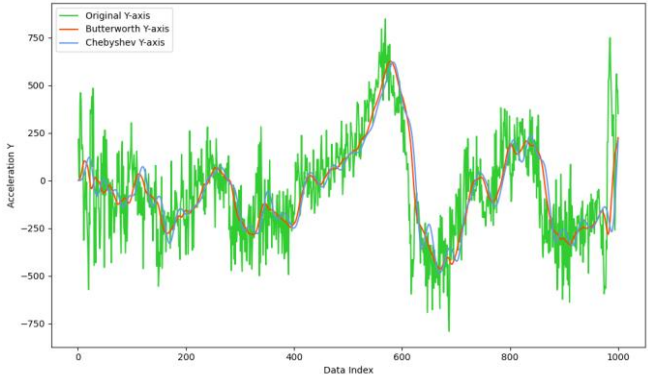

Fig. 10. Acceleration data on y-axis using different filters.

Fig. 11 shows the data captured from GPS where linear interpolation is not applied yet. The blue dot represents location data captured from GPS. As can be seen in the figure mentioned, GPS's ability to capture data is rather abstract. The GPS data is not continuous. It is very dependent on the signal availability. It can cause inconsistency in capturing ability. On the other hand, Fig. 12 is an improvement from the previous figure. The blue line represents the trajectory captured from the UAV location after linear interpolation is implemented. As can be seen in the figure mentioned, UAV movement is continuous and very smooth.

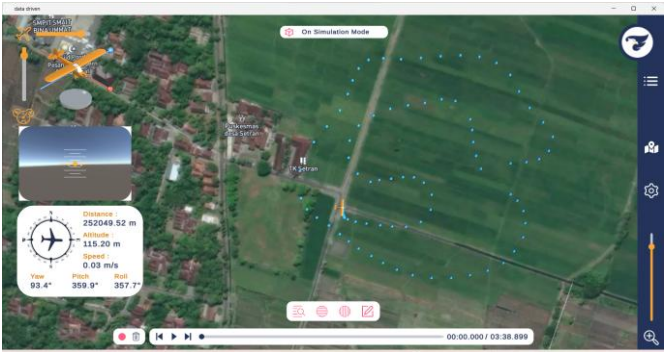

Fig. 11. Location plotted before linear interpolation.

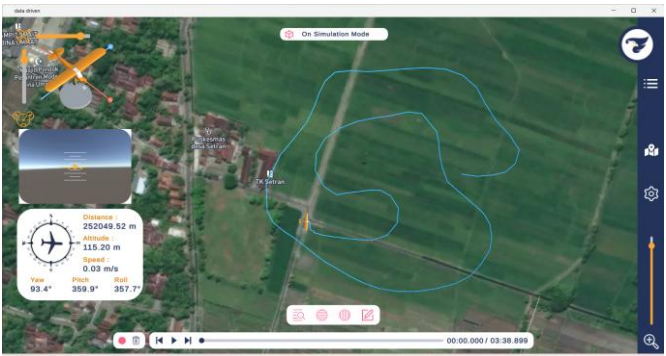

Fig. 12. Location plotted after linear inteprolation.

Fig. 13 is a plot of latitude and longitude data captured from GPS. There are three different data represented with three different colored lines. The blue line represents raw latitude and longitude purely from GPS. The non-continuous spikes indicate that there are unfilled gaps between data. The green line represents interpolated latitude and longitude. This line is much smoother than the raw data because linear interpolation fills the gaps between each data. This interpolated latitude and longitude data is used as the reference of the real-time complemented GPS-accelerometer data denoted by the orange line.

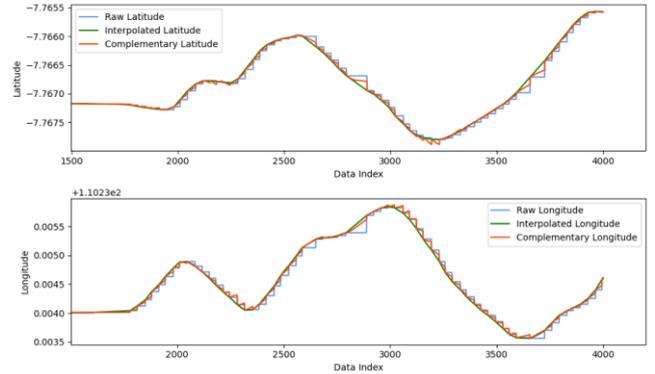

Fig. 13. Latitude and longitude plot.

The complementary filter needs fine-tuning to ensure that the weight from acceleration and GPS data is well complementing each other. After further testing, increasing the weight of α (the weight factor of velocity from acceleration integral) resulted in a spikier graph. On the other hand, increasing the weight of β (the weight factor of velocity integral) resulted in a more delayed graph. Table 1 shows how different α and β impact the latitude and longitude error compared to the interpolated latitude and longitude.

TABLE I. IMPACT OF DIFFERENT ALPHA AND BETA

| α | β | Latitude Error (meter) | Longitude Error (meter) |
|---|---|---|---|
| 0.1 | 0.1 | 2.036 | 2.652 |
| 0.1 | 0.5 | 4.412 | 5.497 |
| 0.1 | 0.9 | 12.866 | 19.465 |
| 0.5 | 0.1 | 3.896 | 4.729 |
| 0.5 | 0.5 | 7.477 | 8.277 |
| 0.5 | 0.9 | 20.816 | 24.115 |
| 0.9 | 0.1 | 6.191 | 7.179 |
| 0.9 | 0.5 | 11.372 | 11.957 |
| 0.9 | 0.9 | 30.411 | 31.506 |

From the data above, it is clear that lower α and β give less errors and thus lead to more precise modeling. Hence, 0.1 is chosen as the value for α and β because it gives lowest error. The graph for complemented latitude and longitude with α and β values of 0.1 is shown in the Fig. 13 presented earlier.

## VI. CONCLUSION

The developed data-driven modeling software successfully provides precise and stable orientation and position data processed from the sensor fusion and data filtering algorithms explained in this paper. For the complementary filter used for position processing from GPS and acceleration data, it has been observed that minimizing the value of α and β results in reduced errors. Additionally, the concept of complementing the first integral of acceleration with the processed vector velocity calculated from scalar speed GPS data has been proven to work well. Regarding the orientation, the inclusion of compass data from the magnetometer has successfully mitigated the error accumulation on the yaw axis. The utilization of quaternion representation has been proven to give a way to model orientation without a gimbal lock problem. Overall, the data shows the software's robust performance, affirming the embedded algorithms' effectiveness.

ACKNOWLEDGMENT

Authors would like to express their sincere gratitude to all Sayakawidya Research and Development Team members—including Fahmi Akmal Zain, Okasah Rofi Izzatik, Kelvin Kurniawan, Adriyan Christhofer Sitanggang, Lauhul Afiat Kahfi, and Leonardo Ginting—who have contributed to the successful completion of this paper. Additionally, authors acknowledge the valuable input received from the reviewers.